\documentclass[10pt,twocolumn,letterpaper]{article}

\usepackage{iccv}
\usepackage{times}
\usepackage{epsfig}
\usepackage{graphicx}
\usepackage{amsmath}
\usepackage{amssymb}

\usepackage{multirow}
\usepackage{cite}
\usepackage{subcaption}
\usepackage{url}
\usepackage{amsthm}
\usepackage{bm}

\theoremstyle{definition}

\iccvfinalcopy 

\def\iccvPaperID{3165} 

\ificcvfinal\fi
\begin{document}

\title{Gyroscope-aided Relative Pose Estimation for Rolling Shutter Cameras}
\author{Chang-Ryeol Lee\\
GIST\\ 
{\tt\small crlee@gist.ac.kr}
\and
Ju Hong Yoon\\
KETI\\
{\tt\small jhyoon@keti.re.kr}
\and
Min-Gyu Park\\
KETI\\
{\tt\small mpark@keti.re.kr}
\and
Kuk-Jin Yoon\\
KAIST\\
{\tt\small kjyoon@kaist.ac.kr}
}

\maketitle

\begin{abstract}
The rolling shutter camera has received great attention due to its low cost imaging capability, however, the estimation of relative pose between rolling shutter cameras still remains a difficult problem owing to its line-by-line image capturing characteristics. 
To alleviate this problem, we exploit gyroscope measurements, angular velocity, along with image measurement to compute the relative pose between rolling shutter cameras. 
The gyroscope measurements provide the information about instantaneous motion that causes the rolling shutter distortion. 
Having gyroscope measurements in one hand, we simplify the relative pose estimation problem and find a minimal solution for the problem based on the Grobner basis polynomial solver.  
The proposed method requires only five points to compute relative pose between rolling shutter cameras, whereas previous methods  require 20 or 44 corresponding points for linear and uniform rolling shutter geometry models, respectively.  
Experimental results on synthetic and real data verify the superiority of the proposed method over existing relative pose estimation methods.
\end{abstract}

\section{Introduction}\label{sec:intro}

\begin{figure}[t]
    \vspace{10pt}
	\centering	
	\begin{subfigure}[t]{0.99\linewidth}
		\includegraphics[width=\linewidth, height=3cm]{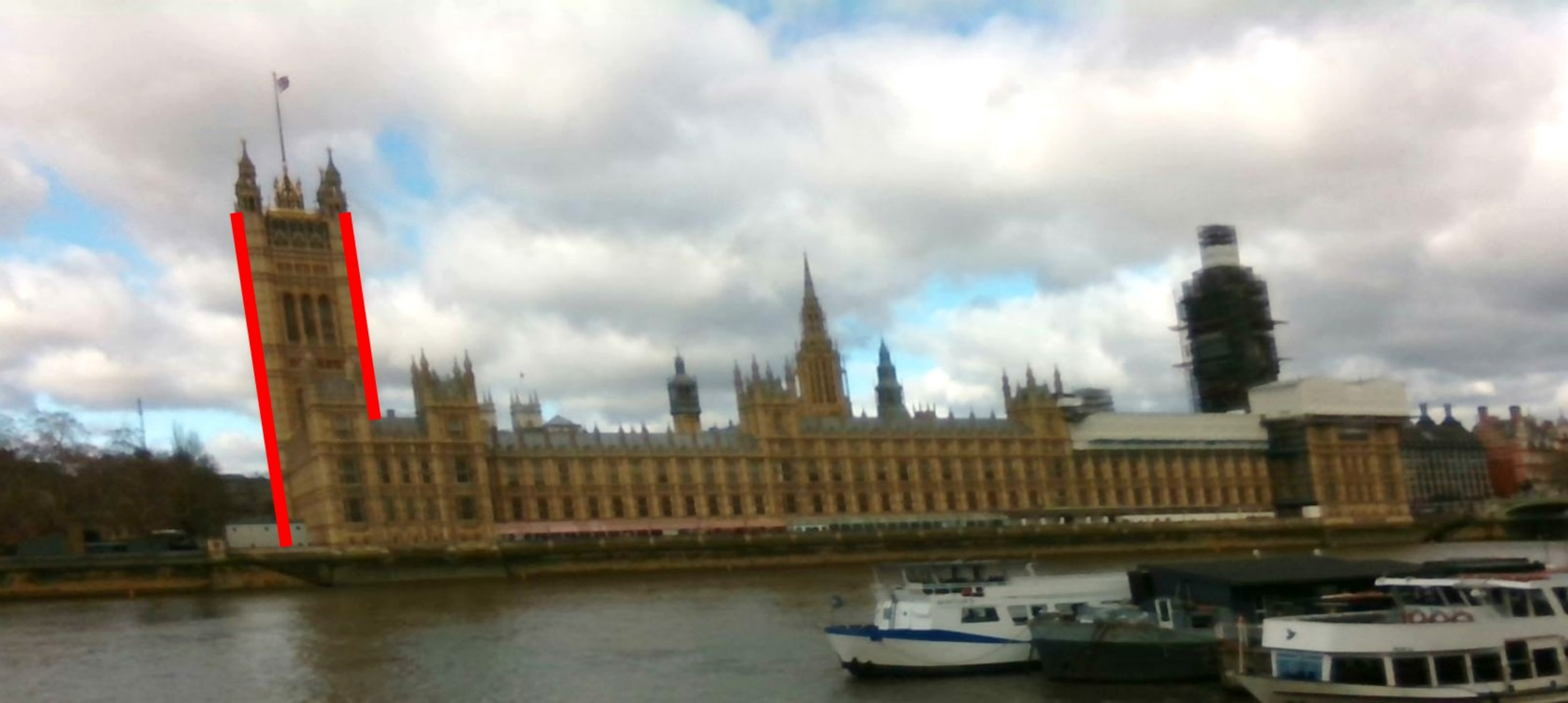}
		\caption{The tower leans towards the left direction.}
	\end{subfigure} \\
	\vspace{2mm}
	\begin{subfigure}[t]{0.99\linewidth}
		\includegraphics[width=\linewidth, height=3cm]{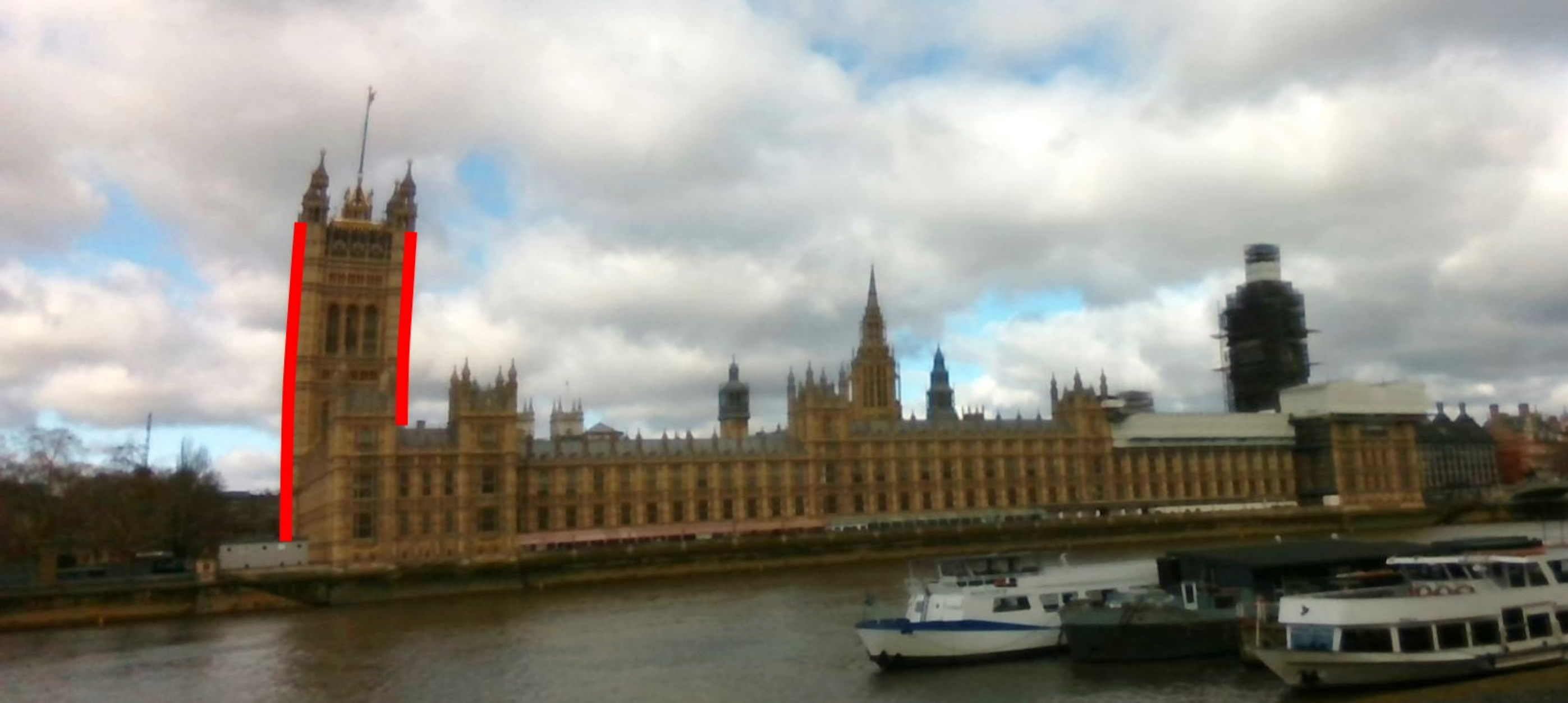}
		\caption{The tower bends to the right direction.}
	\end{subfigure}
	\caption{Examples of distorted images from a rolling shutter camera. The images were captured by a hand-held rolling shutter camera undergoing arbitrary motion.}
	\label{fig:degeneracy_intro} 
	\vspace{-2mm}
\end{figure}

Rolling shutter cameras have become popular due to their low-cost imaging capabilities.
However, their line-by-line image capturing nature causes undesirable artifacts as shown in Fig.~\ref{fig:degeneracy_intro}. 
These artifacts can be critical to geometric vision applications such as structure-from-motion (SfM), simultaneous localization and mapping (SLAM), and dense 3D reconstruction~\cite{Albl:ECCV:2016}.
Therefore, a numerous studies have tackled this problem over the last decade~\cite{Hedborg:ICCVW:2011,Hedborg:CVPR:2012,Saurer:ICCV:2013,Guo:RSS:2014,Kim:ICRA:2016,Albl:CVPR:2015,Albl:ECCV:2016,Dai:CVPR:2016}. 
%

Most existing methods were designed for video applications and the majority of them take a temporal interpolation approach to estimate camera poses~\cite{Hedborg:ICCVW:2011,Hedborg:CVPR:2012,Guo:RSS:2014,Kim:ICRA:2016}. 
They handle the distortion of rolling shutter cameras through nonlinear optimization with a large number of variables and the images from high frame-rate videos. 
Recently, relative pose estimation problems using unordered still images captured by rolling shutter cameras have been also studied~\cite{Albl:ECCV:2016, Dai:CVPR:2016}. 
Albl \etal~~\cite{Albl:CVPR:2015} proposed an absolute pose estimation algorithm for rolling shutter cameras given six pairs of 2D-3D correspondences.  
Besides, they analyzed the degeneracy in rolling shutter structure-from-motion, which happens when the readout directions are close to parallel, and they alleviated the degenerate situation by differentiating readout directions~\cite{Albl:ECCV:2016}. 
In addition, Dai \etal~\cite{Dai:CVPR:2016} proposed linear and nonlinear algorithms for linear and uniform rolling shutter geometry models to solve the rolling shutter relative pose (RSRP) problem. These algorithms require at least 20 and 44 corresponding points for the linear and uniform models on rolling shutter geometry, respectively. 
However, these algorithms are sensitive to outliers and time-consuming due to the  examination of a huge number of hypotheses.
Although they also proposed a  nonlinear solver that utilizes 11 and 17 points for the linear and uniform models respectively, the approach is also computationally demanding because of its nonlinear least square optimization procedure, which is performed at each RANSAC iteration. 
Moreover, the cost function for nonlinear optimization has a large number of unknown parameters and this leads to a local minimum solution rather than a correct solution (global minimum solution). 
One practical approach to alleviate this problem is to lower the degree-of-freedom (DOF) of the problem. 
In this sense, several studies~\cite{Fraundorfer:ECCV:2010, Lee:CVPR:2014, Albl:CVPR:2016, Ovren:ICRA:2015, Karpenko:Tech:2011} adopted inertial measurements to lessen the DOF. 
They utilize the orientation of the gravitational acceleration, which is also called as vertical direction or gravity direction. 
Since the vertical direction is directly related to the inclination of the sensors, they~\cite{Fraundorfer:ECCV:2010, Lee:CVPR:2014} solved the relative pose problem for global shutter cameras with two known angles. 
Albl~\cite{Albl:CVPR:2016} also exploited the vertical direction for the absolute pose estimation of rolling shutter cameras. However, the vertical direction from an IMU is likely to be corrupted by several factors such as sensor noise as they pointed out in the paper, and therefore, additional refinements are required in general. 
On the other hand, some researchers~\cite{Ovren:ICRA:2015, Karpenko:Tech:2011} tried to utilize gyroscope measurements in order to rectify the distorted images for video stabilization.
They also proposed self-calibration, which estimates biases of a gyroscope and the time offset between a gyroscope and a camera, from an image sequence and the corresponding gyroscope measurements.

In this paper, we take one step further with the aid of 3-DOF gyroscope measurements indicating the angular velocity to effectively solve the rolling shutter relative pose (RSRP) problem. 
The angular velocity measurements can be exploited for general situations, \eg a hand-held rolling shutter camera undergoing arbitrary motion. 
Moreover, the angular velocity measurements directly provide the information about instantaneous motion that causes rolling shutter distortion.
From this perspective, it would be also possible to use gravity measurements along with the angular velocity measurements since most of off-the-shelf IMUs provide those measurements together.  %
However, we do not use the gravity measurements because the gravity measurements tend to be very noisy as pointed out in \cite{Albl:CVPR:2016}, and we solve the RSRP problem with only the angular velocity measurements. 
In addition, unlike~\cite{Dai:CVPR:2016}, our work is based on the angular model for rolling shutter camera geometry assuming that instantaneous linear velocities are zero because hand-held camera motion does not contain high linear velocity causing severe distortion in general.
It not only simplifies the problem further but also improves the numerical stability on translation estimation. 
Then, we find the minimal solution for the simplified problems by using the Grobner basis polynomial equation solver.
Note that the proposed algorithms require only five corresponding points for the RSRP problem as the five-point algorithm~\cite{Nister:PAMI:2004} for global shutter cameras. 
In addition, we propose a nonlinear refinement scheme for the estimates obtained from the proposed closed-form method.
The experiments on synthetic and real data show the superiority of the proposed method. We further analyze the performance of the proposed algorithms under various instantaneous motions and noises. 
The real and the synthetic data experiments with non-zero linear velocities substantiate that the proposed methods are effective even under the non-zero linear velocities although we do not consider the linear velocity in our rolling shutter geometry.

\section{Related Work}
\label{sec:related_works}

The rolling shutter camera was addressed in the Perspective-n-Point (PnP) problem to estimate the camera pose with given 2D projections corresponding to 3D point clouds. 
Ait-Aider \etal~\cite{Ait:ECCV:2006} proposed to estimate the pose and the speed of fast-moving objects in a single image with a given 2D-3D matching, and proposed nonlinear and linear models for non-planar and planar objects.
Magerand \etal~\cite{Magerand:ECCV:2012} extended this study by suggesting a polynomial uniform rolling shutter geometry model and solved the problem through the constrained global optimization.
Albl \etal~\cite{Albl:CVPR:2015} proposed a new method to estimate the camera pose with 6 points based on the double linearized rolling shutter camera model.

Besides, SfM and SLAM based on a monocular rolling shutter camera have been intensively studied taking into account the rolling shutter effects. 
Klein and Murray~\cite{Klein:ISMAR:2009} used the constant velocity model in the SLAM framework to predict and correct the rolling shutter distortion occurring in the next frame.
Hedborg \etal~\cite{Hedborg:ICCVW:2011,Hedborg:CVPR:2012} applied the rolling shutter camera projection model to the non-linear optimization step of SfM, \ie bundle adjustment. 
Their key idea is to exploit temporal continuity of the camera motion on the video input to deal with the rolling shutter distortion.
Saurer \etal~\cite{Saurer:ICCV:2013} also considered the rolling shutter distortion in dense 3D reconstruction and they also proposed a minimal solution for absolute pose estimation problem of rolling shutter cameras~\cite{Saurer:IROS:2015}.
Albl \etal~\cite{Albl:ECCV:2016} analyzed the degeneracy of the rolling shutter SfM and suggested how to avoid the degeneracy when shooting videos.
Recently, Ito and Okatani~\cite{Ito:CVPR:2017} derived the degeneracy of rolling shutter SfM as the general expression through a self-calibration-based approach. 
Zhuang \etal~\cite{Zhuang:ICCV:2017} proposed a constant acceleration model for relative pose estimation and image rectification in two consecutive images.

On the other hand, the methods to utilize an inertial measurement unit (IMU) to deal with the rolling shutter distortion in visual odometry (VO) and SLAM have been also studied.
Jia and Evans~\cite{Jia:MMSP:2012} proposed a method to estimate the camera orientation using gyroscope measurements and to correct the rolling shutter distortion of the image.
Guo~\etal~\cite{Guo:RSS:2014} applied a rolling shutter camera projection model to the visual-inertial odometry framework that uses IMUs and cameras to estimate egomotion.
They estimated the readout time of the rolling shutter camera as well as the time delay between the IMU and the camera.
Albl~\etal~\cite{Albl:CVPR:2016} proposed a method to improve the speed and accuracy of the method in \cite{Albl:CVPR:2015} using the gravity measurements obtained from the IMU.
In addition, IMUs are also used to solve conventional relative or absolute pose estimation problems for global shutter cameras.
It has been studied to estimate relative pose with the partially known orientation angle between two cameras  \cite{Fraundorfer:ECCV:2010,Li:IROS:2013}, or with known vertical direction  \cite{Kukelova:ACCV:2011,Lee:CVPR:2014}.

The relative pose estimation is a fundamental problem and of importance in SfM and SLAM.
To the best of our knowledge, this is the first work to exploit gyroscope measurements for the relative pose estimation of the rolling shutter cameras.

\section{The Proposed Method} \label{sec:proposed}

In this section, we define and formulate the gyroscope-aided RSRP problem, which simplifies the RSRP problem having 11 DOF as described in Eqs \eqref{eq:rolling_shutter_epipolar_geometry}-\eqref{eq:three_rolling_shutter_rotation_matrix}.
With the gyroscope measurements, 11 DOF of the RSRP problem is decreased to 5 DOF because we utilize  known three-dimensional angular velocities from gyroscopes in two different view points. 
Then, we solve the problems using the Grobner Basis (GB) method in order to obtain a closed form solution. 
To apply the GB method, we simplify the RSRP problem with the triple-linearized model for rotation parameters.
In the final step, the estimated parameters from the GB method are used as initial values for the nonlinear refinement.

\subsection{Rolling Shutter Relative Pose Problem} \label{sec:problem}
In this paper, we use the angular model for rolling shutter geometry as in~\cite{Ovren:ICRA:2015, Karpenko:Tech:2011}, in which the rolling shutter distortion is represented by the instant angular velocity not the instant linear velocity because the angular velocity dominates the image distortion of rolling shutter cameras in general.
Indeed, the rolling shutter image distortion usually becomes severe when captured scenes are close to a camera and the camera moves very fast. In such cases, it is difficult to get the proper number of corresponding points between two images for relative pose estimation, and the linear velocity is a very small value in comparison with the fast camera motion.
Therefore, considering linear velocity is practically not effective unlike the absolute pose problems in~\cite{Saurer:IROS:2015, Albl:CVPR:2015, Albl:CVPR:2016}, and this is validated in experiments.
Moreover, the linear velocity increases the DOF of the RSRP problem and makes the relative translation estimation much more difficult and unstable.
For those reasons, we omit the linear velocity from the rolling shutter geometry in this paper.

Then, the geometric relations between a pair of corresponding points, $\mathbf{m}_i$ and $\mathbf{m}_j$, in the normalized camera coordinate is formulated by using the epipolar constraint as
\begin{equation}
\mathbf{m}_j^{\top} \mathbf{E}_{v_i v_j} \mathbf{m}_i = 0,
\label{eq:rolling_shutter_epipolar_geometry}
\end{equation}
Here, $i$ and $j$ indicate a frame index or a camera index for different cameras throughout the paper. $v_i$ and $v_j$ are a row index of images. The rolling shutter essential matrix $\mathbf{E}_{v_i v_j}$ is defined as 
\begin{equation}
\mathbf{E}_{v_i v_j} = \lfloor \mathbf{t} \rfloor_{\times} \mathbf{R}_{v_i v_j}.
\label{eq:rolling_shutter_essential_matrix}
\end{equation}
Here, $\mathbf{t} \in \mathbb{R}^{3}$ a relative translation vector and the rotation matrix $\mathbf{R}$ is defined as
\begin{equation}
\begin{split}
\mathbf{R}_{v_i v_j} & = {\mathbf{R}\left( v_j \lambda \mathbf{w}_j \right)}^{\top}  \mathbf{R}(\mathbf{a}) \mathbf{R} \left( v_i \lambda \mathbf{w}_i \right),
\end{split}
\label{eq:three_rolling_shutter_rotation_matrix}
\end{equation}
where $\mathbf{R}(\cdot) \in \text{SO}(3)$ denotes a rotation matrix, $\mathbf{w} \in \mathbb{R}^{3}$ denotes an instant angular velocity vector, $\mathbf{a} \in \mathbb{R}^{3}$ represents the relative rotation between two cameras, and $\lambda$ is a readout time.
The distortion from the instant angular velocities is expressed as $\mathbf{R}(\cdot)$ with angular velocity vectors.
Solving the rolling shutter relative pose problem is identical to computing $\mathbf{R}$ and $\mathbf{t}$ from the rolling shutter essential matrix in~ Eq. \eqref{eq:rolling_shutter_essential_matrix}.

\begin{figure}[t]
	\centering	
	\includegraphics[width=\linewidth]{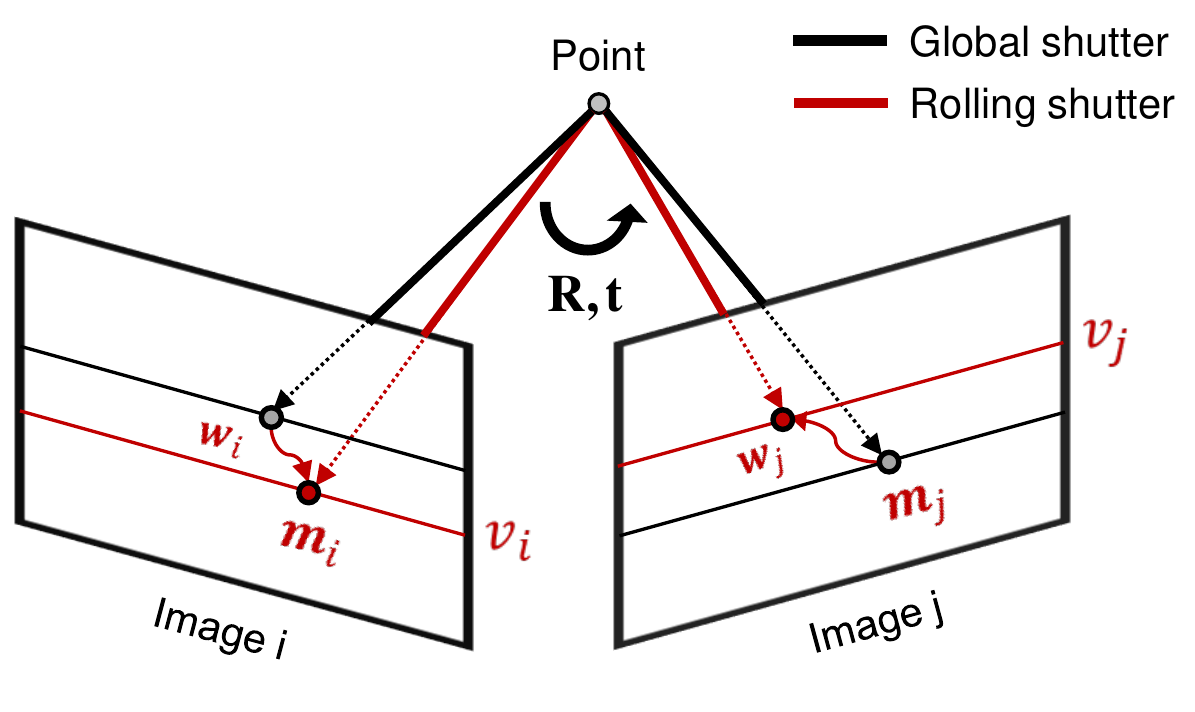}	
	\caption{Two-view geometry of a rolling shutter camera. Camera motion is denoted by relative rotation $\mathbf{R}$ and translation $\mathbf{t}$. $\mathbf{m}_i$ and $\mathbf{m}_j$ are corresponding points. The rolling shutter distortion is described by angular $\mathbf{w}$. $v$ represents a row index.}		
	\label{fig:rs_two_view_geometry} 
\end{figure}

The difference between a rolling shutter camera and a global shutter camera is depicted in Fig.~\ref{fig:rs_two_view_geometry}; the rolling shutter essential matrix has 11 DOF ($dim(\mathbf{R})$ = 3, $dim(\mathbf{w}_i)$ = 3, $dim(\mathbf{w}_j)$ = 3, $dim(\mathbf{t})$ = 2).
Since the the relative translation is up to scale, its DOF is two.

Here, we summarize our system setup and the RSRP problem with a gyroscope.
We have a calibrated rolling shutter camera--gyroscope system, therefore, we know intrinsics of a camera and a gyroscope, extrinsics between a camera and a gyroscope, and rolling shutter readout times.
Then, the inputs of the problem are gyroscope measurements (\ie angular velocity) and corresponding image points in two images.
The output is the relative pose between the two rolling shutter cameras, \ie the rotation $\mathbf{R}$ and translation $\mathbf{t}$.

\subsection{Angular RSRP with Gyroscope Measurements} \label{subsec:grsrp}

The angular velocity measurements from the gyroscope of the each camera reduce the DOF of the RSRP problem from 11 to five (\ie $\mathbf{R}, \mathbf{t}$) because the angular velocities of the problem are given and 
the relative translation is up to scale.
While it is simpler than the original RSRP problem having 11 DOF, we further simplify the problem in order to obtain feasible solutions for real-world applications. 

We can express rotations as polynomials for the GB method with the Cayley transform~\cite{hazewinkel:BOOK:2013,Albl:CVPR:2015}. 
It has the denominator $K = 1 + x^2 + y^2 + z^2$ from the input angle vector $[x, y, z]^\top\in \mathbb{R}^3$.
It can be easily removed by multiplication of $K$ to the rolling shutter epipolar constraint in Eq.~\eqref{eq:rolling_shutter_epipolar_geometry} in order to obtain pure polynomials for the GB method.
However, it still has second-order monomials on $x$, $y$, and $z$.
Thus, a system to use the Cayley transform for the three rotation matrices in Eq.~\eqref{eq:three_rolling_shutter_rotation_matrix} has a huge number of monomials and the large elimination template from the GB method.
Therefore, since this system based on the Cayley transform is very time-consuming and numerically unstable, we do not directly use the Cayley model.

Instead, we express the rotation matrices from instant angular velocities as the linearized model.
It is a sensible assumption because an image is captured in very short time, commonly 30$ms$ to 50$ms$, and the instant rotation in rolling shutter cameras is typically very small. Under this assumption, the  rotation matrices of cameras can be expressed as
\begin{equation}
    \mathbf{R} ( v  \mathbf{w}^{'}) \approx \mathbf{I} + \lfloor  v \mathbf{w}^{'} \rfloor_{\times}.
    \label{eq:angular_velcoity_rotation_linearized_model}
\end{equation} 
We do substitution $ \lambda \mathbf{w} \xrightarrow{} \mathbf{w}^{'}$ for brevity and reducing the number of multiplication in the estimation process.
As a result, we have the following rotation representation.
\begin{equation}
\mathbf{R}_{v_i v_j}  = \left(\mathbf{I} + v_j \lfloor \mathbf{w}_j^{'} \rfloor_{\times} \right)^{\top}  \mathbf{R}(\mathbf{a}) \left(\mathbf{I} + v_i  \lfloor \mathbf{w}_i^{'} \rfloor_{\times} \right)
\label{eq:linearized_rolling_shutter_rotation_matrix_1}
\end{equation}
where $\lfloor \cdot \rfloor_{\times}$ is the skew symmetric matrix form of a cross product. 

Besides, we simplify the relative rotation matrix involving $\mathbf{a} \in \mathbb{R}^3$, however, it is not a small rotation unlike the rotations from instant motion.
Therefore, we apply the linearization with an initial rotation $\mathbf{R}_0$ as in~\cite{Albl:CVPR:2016}, and obtain  $\mathbf{R} (\mathbf{a})$ as 
\begin{equation}
    \mathbf{R} (\mathbf{a}) \approx \mathbf{R}_0 (\mathbf{I} + \lfloor \mathbf{a} \rfloor_{\times}).
    \label{eq:relative_rotation_linearization}
\end{equation} 
Finally, we obtain the rolling shutter relative rotation matrix as below:
{
\begin{equation}
\begin{split}
& \mathbf{R}_{v_i v_j}  = \\ & \left(\mathbf{I} + v_j \lfloor \mathbf{w}_j^{'} \rfloor_{\times} \right)^{\top}  \mathbf{R}_0 (\mathbf{I} + \lfloor \mathbf{a} \rfloor_{\times}) \left(\mathbf{I} + v_i \lfloor \mathbf{w}_i^{'} \rfloor_{\times} \right).
\end{split}
\label{eq:linearized_rolling_shutter_rotation_matrix_2}
\end{equation}
}

In order to solve the problem, we construct six equations from the rolling shutter essential matrix with five corresponding points and one constraint equation from the scale ambiguity as below:
\begin{equation}
    \left \lVert \mathbf{t} \right \rVert_2 =  1.
    \label{eq:scale_constraint_r_definition}
\end{equation}
We reformulate this scale constraint as a polynomial form for the GB method as 
\begin{equation}
    t_x^2 + t_y^2 + t_z^2 - 1 = 0.
    \label{eq:scale_constraint}
\end{equation}
 
We obtain six equations for six unknowns and, from those equations, we have 20 possible solutions.
By using the automatic generator from~\cite{kukelova:ECCV:2008}, we compute the elimination template that describes which polynomials have to be added to initial equations to obtain Grobner basis and all polynomials required for constructing action matrix.
To get the elimination template, we  multiply all the monomials of the six equations up to 5-DOF, and generate 205 polynomials from 462 monomials. 
After removing unnecessary polynomials and monomials, we obtain a final (205$\times$225)-dimensional elimination template.
Note that we do not need to compute it again once we find the elimination template in a pre-processing stage.

The elimination template, which contains coefficients from input measurements such as corresponding points and gyroscope measurements, is converted to the reduced row echelon form, and the 20$\times$20 action matrix is constructed from the template.
Then, we obtain solutions from eigenvalues and eigenvectors of the action matrix.
We choose one solution among the 20 solutions by removing imaginary solutions.

\subsection{Initialization and Robust Estimation}
\label{sec:initialization}
 
The initial rotation $\mathbf{R}_{0}$ can be obtained from several ways.
Firstly, the commercial IMUs usually contain an accelerometer that provides gravity measurements as well as a gyroscope.
The vertical direction of a camera from gravity measurements (\ie pitch and roll angles) can be used to compute two-dimensional relative rotation between two cameras. 
Also, we can predict the initial pose from a motion model in case of SLAM applications.
In this paper, we obtain the initial rotation $\mathbf{R}_0$ by using the well-known five-point relative pose estimation algorithm~\cite{Nister:PAMI:2004}.
 
To increase robustness against noise and/or outliers, we apply the RANSAC\cite{Fischler:RANSAC} to our method. 
%
In each RANSAC iteration, we choose the solution with the largest number of inliers among 20 solutions. 
We use the rolling shutter epipolar constraint in Eq.~\eqref{eq:rolling_shutter_epipolar_geometry} for the cost function of the RANSAC.
A threshold for determining inliers is set to 0.01 and the number of iteration is set to 200.

\subsection{Nonlinear Refinement} \label{subsec:refinement}

Although the proposed method estimates the relative pose of two rolling shutter cameras accurately, we further enhance the pose accuracy through nonlinear optimization. 
In this step, we use the pose estimation result from the proposed method described in the previous section as an initial relative pose, and further refine the pose through the nonlinear optimization with the inlier points obtained from the RANSAC described in Sec.~\ref{sec:initialization}. 
%
The nonlinear energy minimization is formulated as 
\begin{equation}
\min_{\mathbf{q},\mathbf{t} }{\sum_{uv}{C(\mathbf{q},\mathbf{t}  )^2}}  \ \ \ \text{s.t} \  \left\lVert \mathbf{t} \right\rVert_{2} = 1,
\label{eq:cost_refinement}
\end{equation}
where $\mathbf{t}$ is the translation in Eq. \eqref{eq:scale_constraint_r_definition} and $\mathbf{q} \in \mathbb{R}^{4}$ is the unit-quaternion representing the relative rotation, and it is estimated with local parameterization $\bm{\theta} \in \mathbb{R}^3$ against its tangent plane.
The scale ambiguity constraint is added to the cost function as the form of Eq.~\eqref{eq:scale_constraint} for easy differentiation.  
The cost function $C$ is defined as follows.
\begin{equation}
C(\mathbf{q},\mathbf{t})  = \mathbf{m}_j^{\top} {\lfloor \mathbf{t}  \rfloor_{\times}}\mathbf{R}_{v_i v_j} \mathbf{m}_i, 
\end{equation}
where $\mathbf{R}_{v_i v_j}$ is expressed as 
\begin{equation}
\mathbf{R}_{v_i v_j}  = \left(\mathbf{I} + v_j   \lfloor \mathbf{w}_j^{'} \rfloor_{\times} \right)^{\top}  \mathbf{R}_q \left(\mathbf{I} + v_i  \lfloor \mathbf{w}_i^{'} \rfloor_{\times} \right).
\end{equation}
Here, $\mathbf{R}_q$ is the relative rotation between two cameras transformed from the quaternion $\mathbf{q}$ and it is initialized by the estimates from Section~\ref{subsec:grsrp}.
Through this refinement, approximation errors regarding relative rotation $\mathbf{R}$ in Eq.~\eqref{eq:relative_rotation_linearization} are minimized.

\section{Experimental Results} \label{sec:exerpiemtns}

We evaluate the proposed methods on both synthetic and real datasets.
In synthetic data experiments, we analyze the performance of the proposed methods with various factors such as rolling shutter distortions and measurement noises.
Then, we validate the superiority of the proposed methods in reality with real data.

For evaluation, we compare the basic global shutter five-point algorithm~\cite{Kneip:ICRA:2014,Nister:PAMI:2004} with the proposed rolling shutter algorithm and its refined results.
Then, we also compare the nonlinear solver of \cite{Dai:CVPR:2016} that uses estimates of GS five-point algorithm as an initial value.
Since it does not provide inliers, we use a robust estimator, Cauchy loss function, with all measurements.

We summarize the algorithms used in the experiments as follows.
\vspace{-1mm}
\begin{itemize}
    \setlength{\itemsep}{0pt}
    \setlength{\parskip}{0pt}
    \setlength{\parsep}{0pt}
    \item \textbf{GSRP}: The five-point global shutter relative pose estimation algorithm~\cite{Kneip:ICRA:2014,Nister:PAMI:2004}.
    \item \textbf{NRSRP}: The nonlinear rolling shutter relative pose estimation algorithm~\cite{Dai:CVPR:2016}.
    \item \textbf{G-RSRP}: The proposed rolling shutter relative pose estimation algorithm using gyroscope measurements.
    \item \textbf{G-RSRP+}:  The proposed algorithm (G-RSRP) with the additional nonlinear refinement.
\end{itemize}

The evaluation metric for relative rotation evaluation is defined as
\begin{equation}
    \cos^{-1}((\text{trace}(\mathbf{R}_{\text{gt}}^{-1}\mathbf{R}_{\text{est}})-1)/2) 
\end{equation}
and for relative translation it is defined as
\begin{equation}
    \cos^{-1}(\mathbf{t}_{\text{gt}}^{\top} \mathbf{t}_{\text{est}}) 
\end{equation}
as in \cite{Dai:CVPR:2016}, 
where $\mathbf{t}$ is assumed to be a unit vector since the translation is up to scale in the RSRP problem. 
The relative translation error indicates the geodesic distance between ground truth and estimated translation on a unit sphere.

\subsection{Experiments on Synthetic Data}

In the synthetic data experiments, we perform two kinds of experiments: 1) with different angular and linear velocities and 2) with noises of a camera and a gyroscope.

\textbf{Synthetic data generation: } 
For each evaluation, we randomly generate 3D points, and positions/orientations of two cameras to generate synthetic data.
The standard deviation of the relative rotation and translation are set to 10$^{\circ}$ and 2$m$, respectively.
The number of the 3D points is set to 150 and the they are generated within 60$m$ distance from the cameras.
The rolling shutter readout time $\lambda$ is set to 60$us$.
The image resolution is set to $1920 \times 1080$, focal length is set to $640$ pixel, and radial distortion is not considered in this experiment.
The generated points are projected onto the image plane of each camera with the given intrinsic camera parameters and rolling shutter parameters (angular, linear velocities, and readout time).
The rotation from instant angular velocities is computed with the Cayley transform~\cite{hazewinkel:BOOK:2013}, not a linearized model that is used in the proposed method.
Then, we remove the points that are out of the field of view or have no corresponding points.

\textbf{Experiments: } 
We repeat all the experiments 100 times to obtain statistically meaningful results.
At first, we perform experiments as  increasing instantaneous camera motion in order to analyze the effects of angular velocity and linear velocity.
Here, the angular velocities $\mathbf{w}_1, \mathbf{w}_2$ and linear velocities $\mathbf{d}_1,\mathbf{d}_2$ are randomly generated.
But we manually set the magnitude of them in order to see the effect on different level of velocities. 
We designed two experiments: 1) increasing angular velocity with zero linear velocity and 2) increasing angular and linear velocities together.
Although the proposed method assumes the angular rolling shutter geometry model, we perform the experiments with the non-zero instant linear velocities in order to see the applicability and robustness of the proposed method in reality.
For that reason, we increase the magnitudes of the velocities at five levels.
Then, Gaussian noises are added to the camera and gyroscope measurements.
The standard deviations of the noises are set to 1 $pixel$ and 0.1 $rad/s$.

\begin{figure}[t]
	\begin{subfigure}[b]{0.99\linewidth}
		\includegraphics[width=\linewidth]{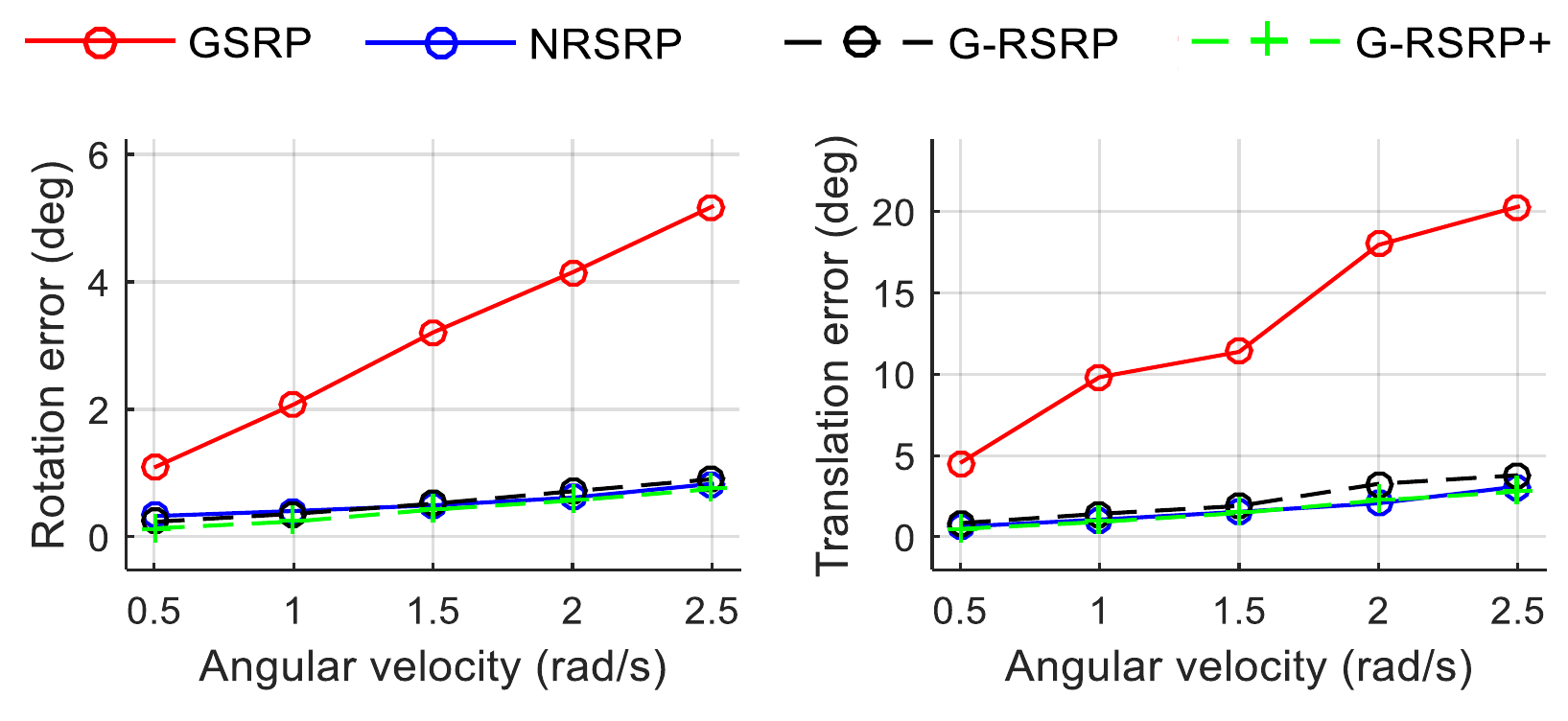}
		\caption{Average}
	\end{subfigure} \vspace{5pt} \\ 
	\begin{subfigure}[b]{0.99\linewidth}
		\includegraphics[width=\linewidth]{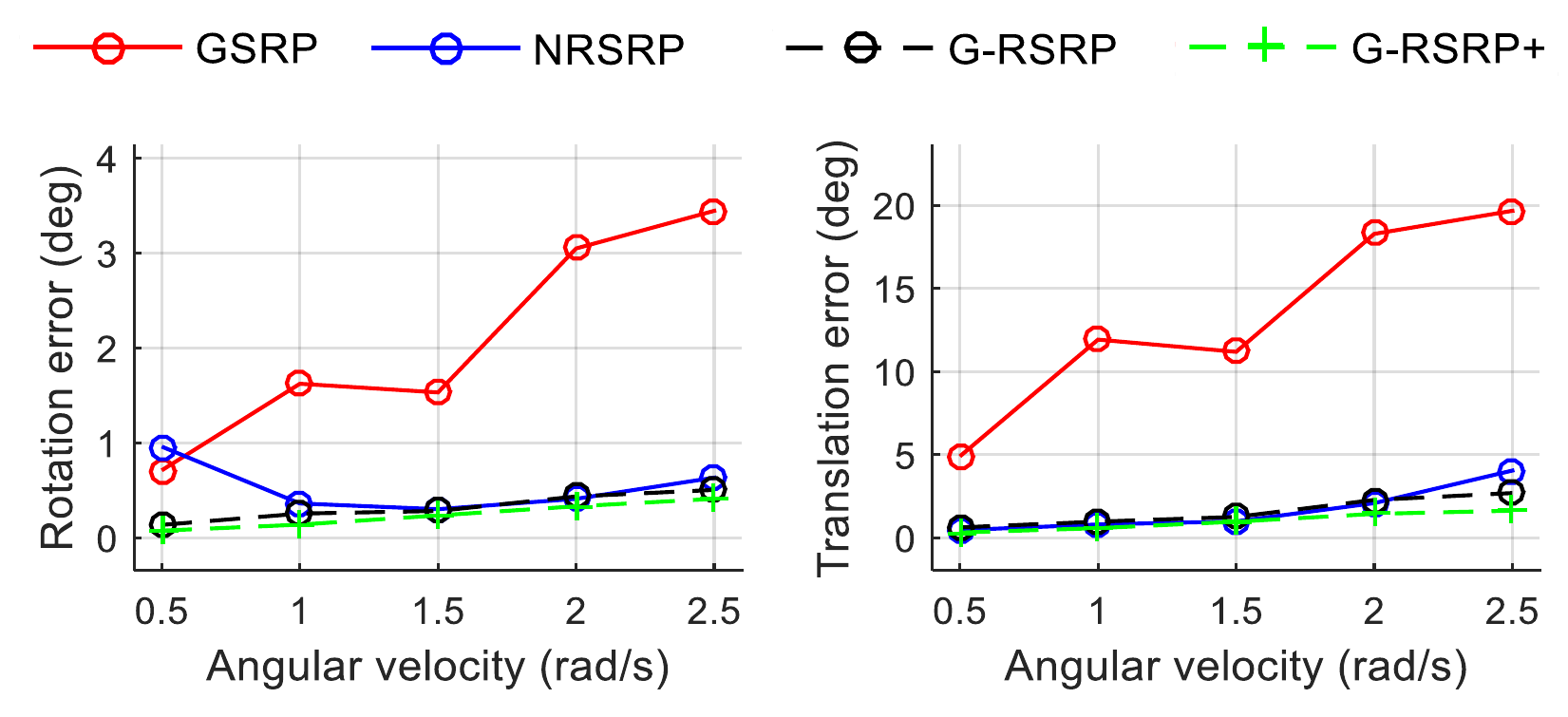}
		\caption{Standard Deviation}
	\end{subfigure}%
	\caption{Performance comparison with different levels of the angular velocity}
	\label{fig:exp_synthetic_rs_w} 
\end{figure}

\begin{figure}[t]
	\begin{subfigure}[b]{0.99\linewidth}
 		\includegraphics[width=\linewidth]{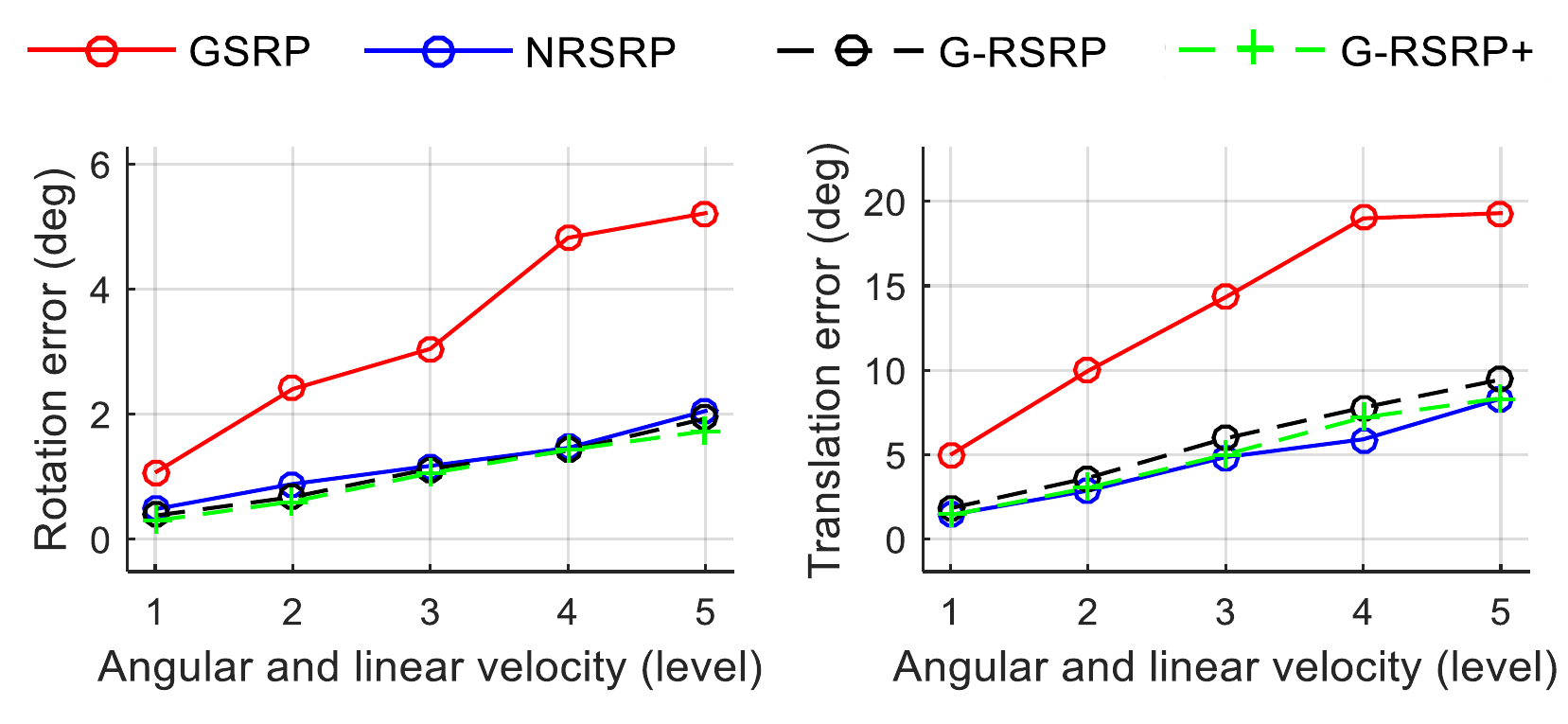}
		\caption{Average}
	\end{subfigure} \vspace{5pt} \\ 
	\begin{subfigure}[b]{0.99\linewidth}
 		\includegraphics[width=\linewidth]{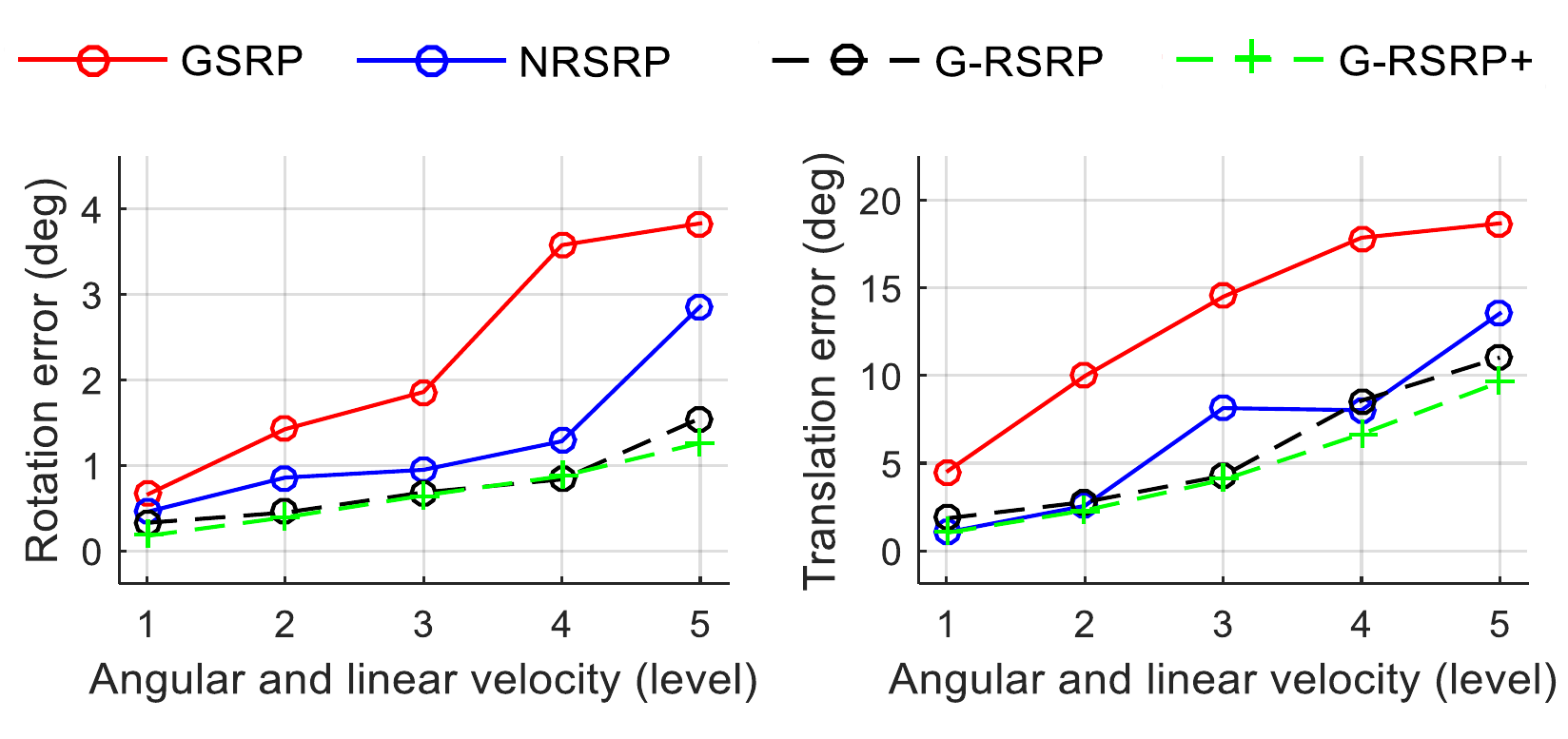}
		\caption{Standard Deviation}
	\end{subfigure}%
	\caption{Performance comparison with different levels of angular and linear velocities}
	\label{fig:exp_synthetic_rs_dw} 
	\vspace{-2mm}
\end{figure}

\begin{figure}[t]
	\center
	\begin{subfigure}[t]{0.99\linewidth}
		\includegraphics[width=\linewidth]{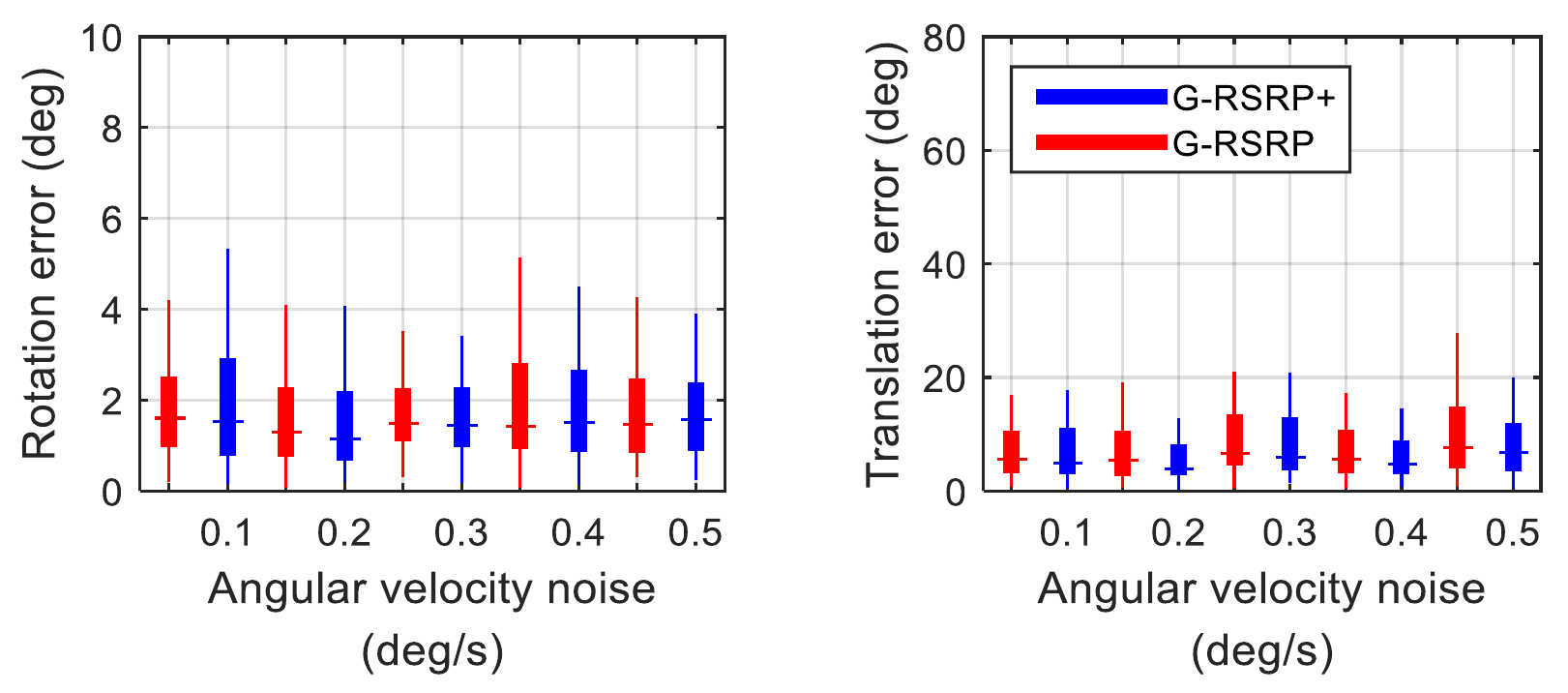}
		\caption{With different levels of angular velocity noises}
		\label{fig:exp_synthetic_w_noise} 
	\end{subfigure} \\ \vspace{5pt}
	\begin{subfigure}[t]{0.99\linewidth}
		\includegraphics[width=\linewidth]{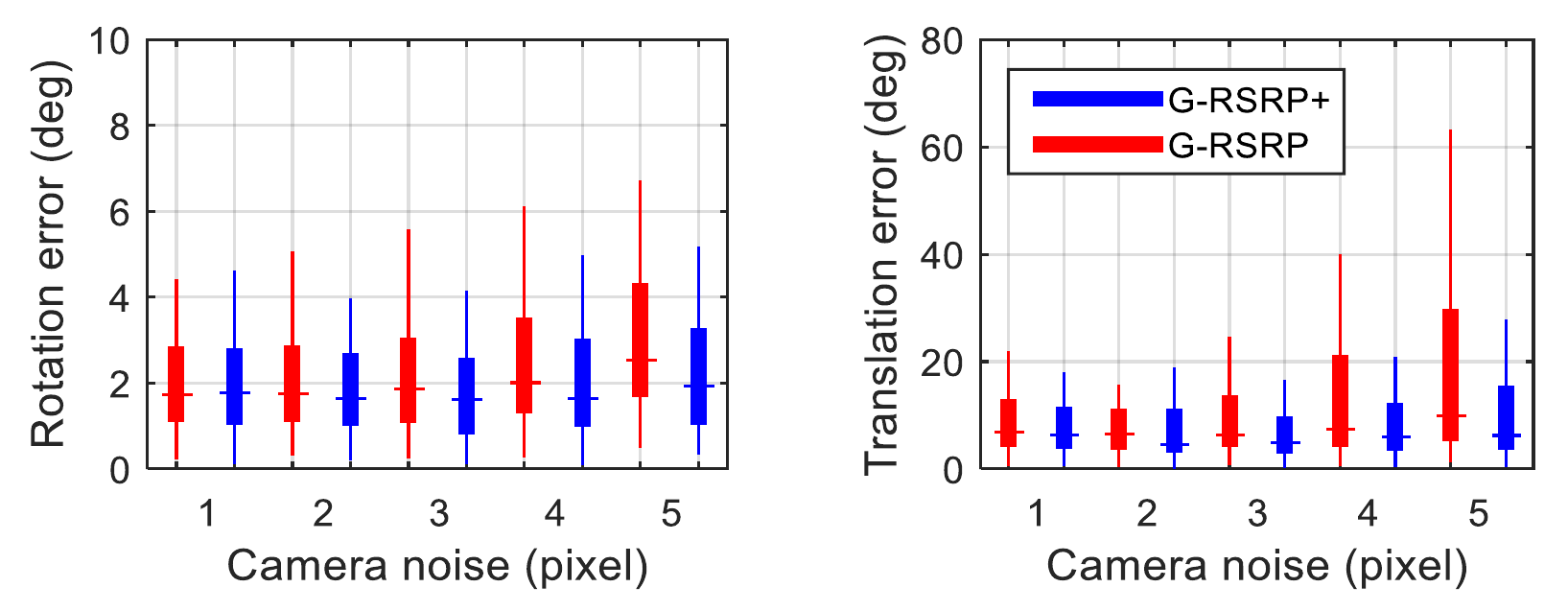}
		\caption{With different levels of camera noises}
		\label{fig:exp_synthetic_camera_noise} 
	\end{subfigure}
	\caption{Performance comparison with different levels of noises}
	\label{fig:exp_synthetic_noise}  	
	\vspace{-4mm}
\end{figure}

For the experiment on different levels of angular velocity with zero linear velocity, the magnitude of the angular velocity is increased from $0.5$ to $2.5 \ rad/s$ by $0.5 \ rad/s$.
Figure~\ref{fig:exp_synthetic_rs_w} shows that the average rotation and translation errors of GSRP~\cite{Nister:PAMI:2004} rapidly increase compared to NRSRP, G-RSRP, G-RSRP+ as the angular velocity increases. 
G-RSRP, G-RSRP+, and RSRP maintain rotation errors less than 1$^{\circ}$ and translation errors less then 5$^{\circ}$ on average even though the angular velocity increases.
Besides, the standard deviation of the rotation and translation errors from G-RSRP, G-RSRP+, and NRSRP are less than 1$^{\circ}$ and 5$^{\circ}$, respectively.
It means that they produce stable estimates. 
Besides, the standard deviation of the errors from G-RSRP+ is less that of G-RSRP.
It indicates that the nonlinear refinements truly improve the stability of the estimation especially under large angular velocities. 
However, NRSRP generates rather large standard deviation at the 0.5 $rad/s$ angular velocity because the nonlinear solver could be fall into local minima in the case that the initial value is not reliable.

\begin{figure*}[t]
	\centering	 
	\begin{subfigure}[t]{0.33\linewidth}
		\includegraphics[width=\linewidth]{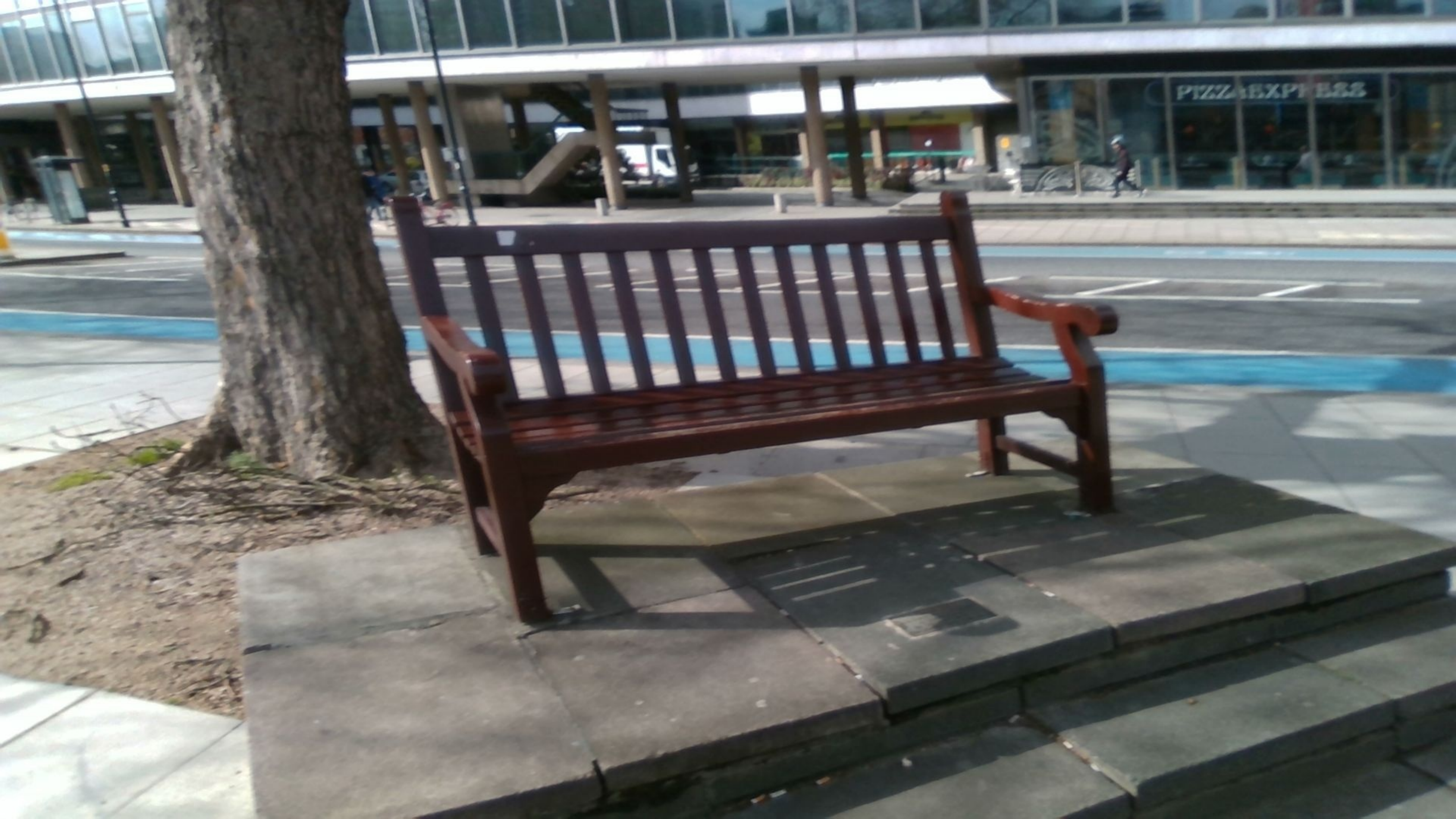}
	\end{subfigure} 
	\begin{subfigure}[t]{0.33\linewidth}
		\includegraphics[width=\linewidth]{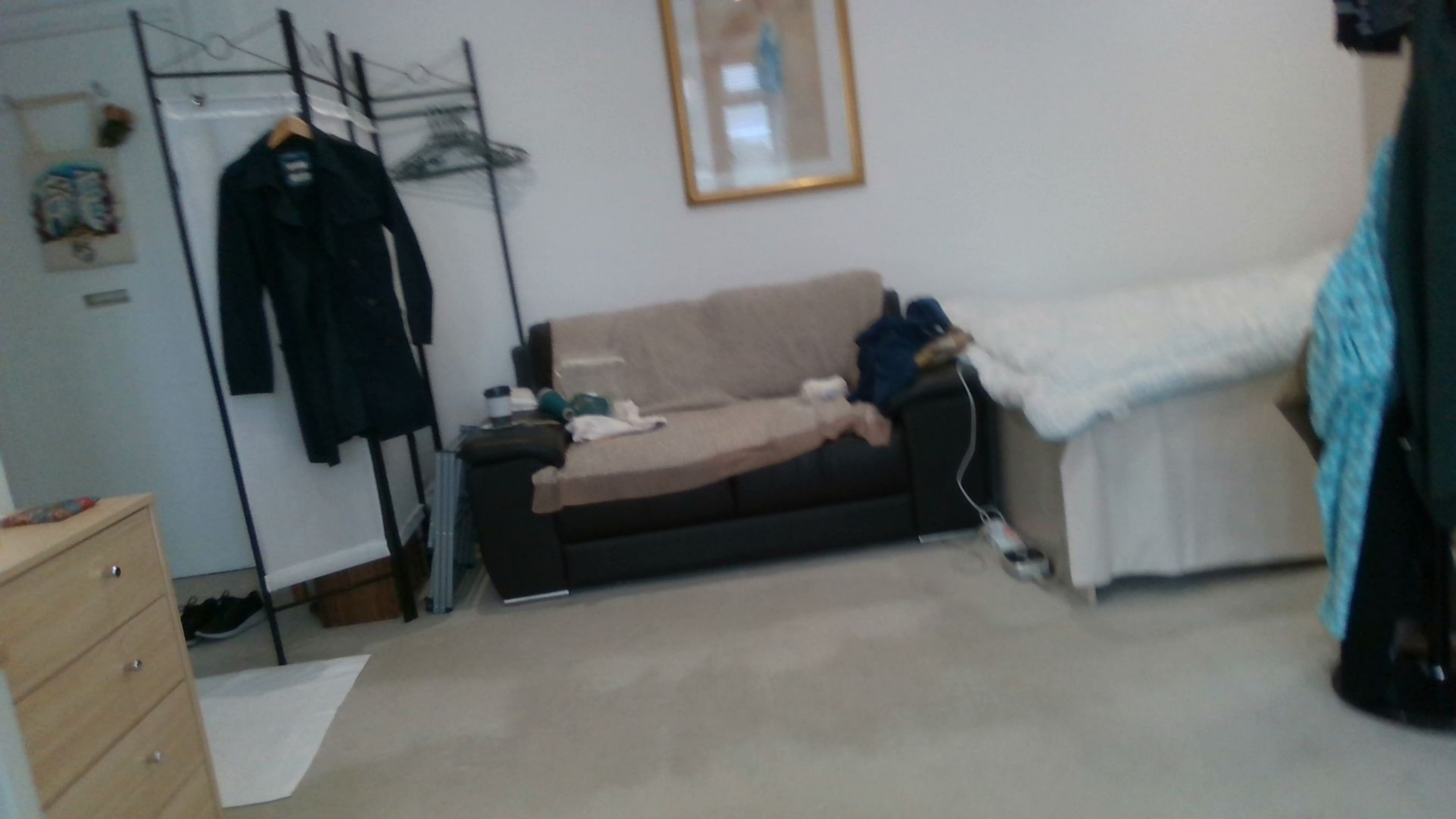}
	\end{subfigure} 
	\begin{subfigure}[t]{0.33\linewidth}
		\includegraphics[width=\linewidth]{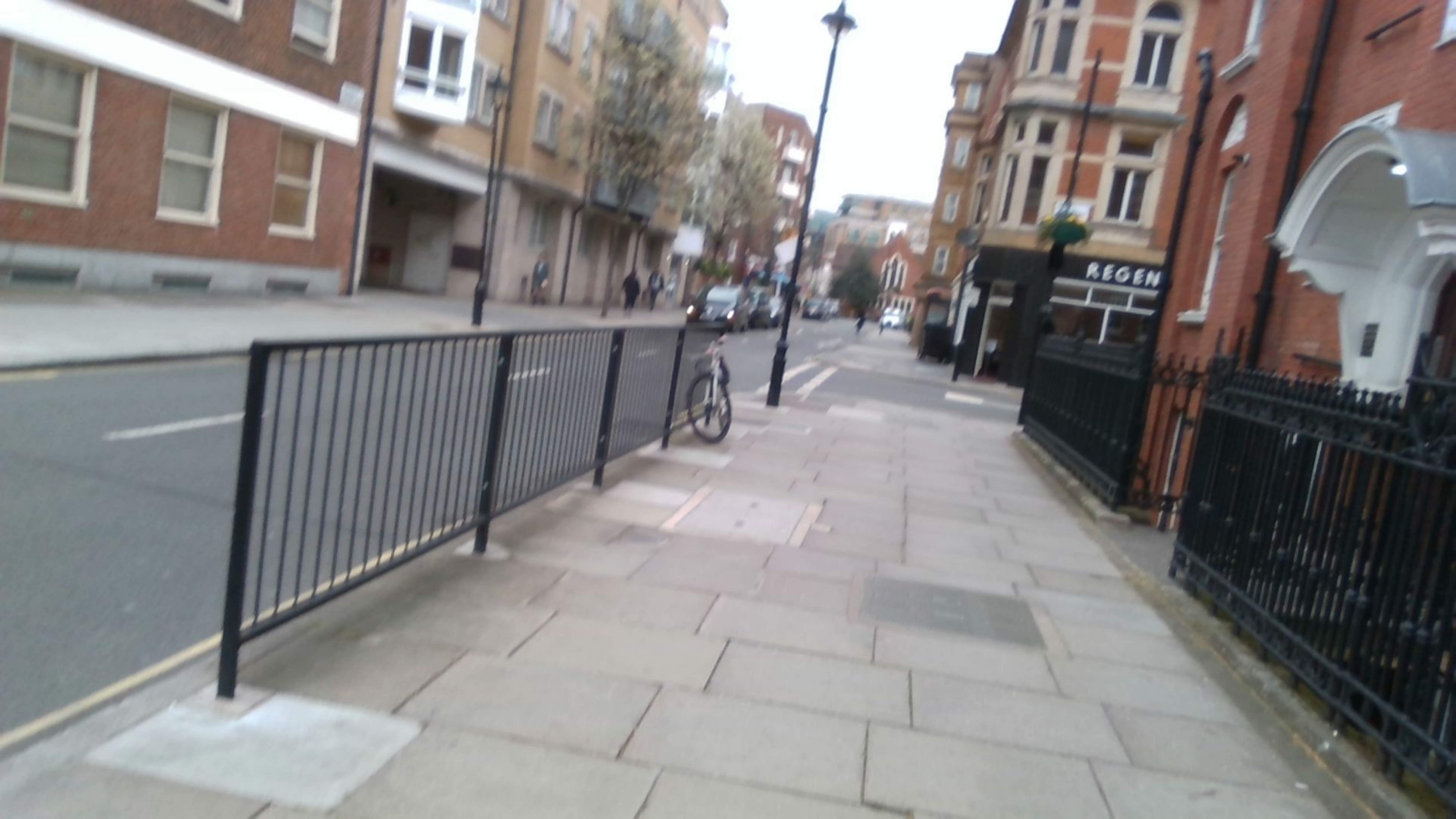}
	\end{subfigure}	\\ \vspace{1mm}
	\begin{subfigure}[t]{0.33\linewidth}
		\includegraphics[width=\linewidth, height=2cm]{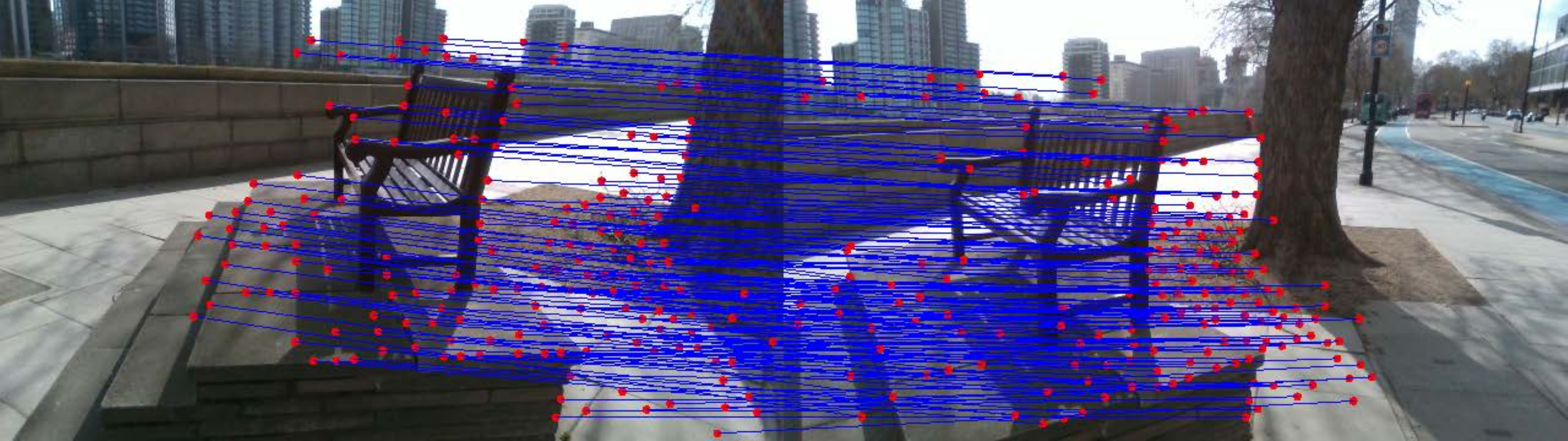}
		\caption{Bench}
	\end{subfigure}  
	\begin{subfigure}[t]{0.33\linewidth}
		\includegraphics[width=\linewidth, height=2cm]{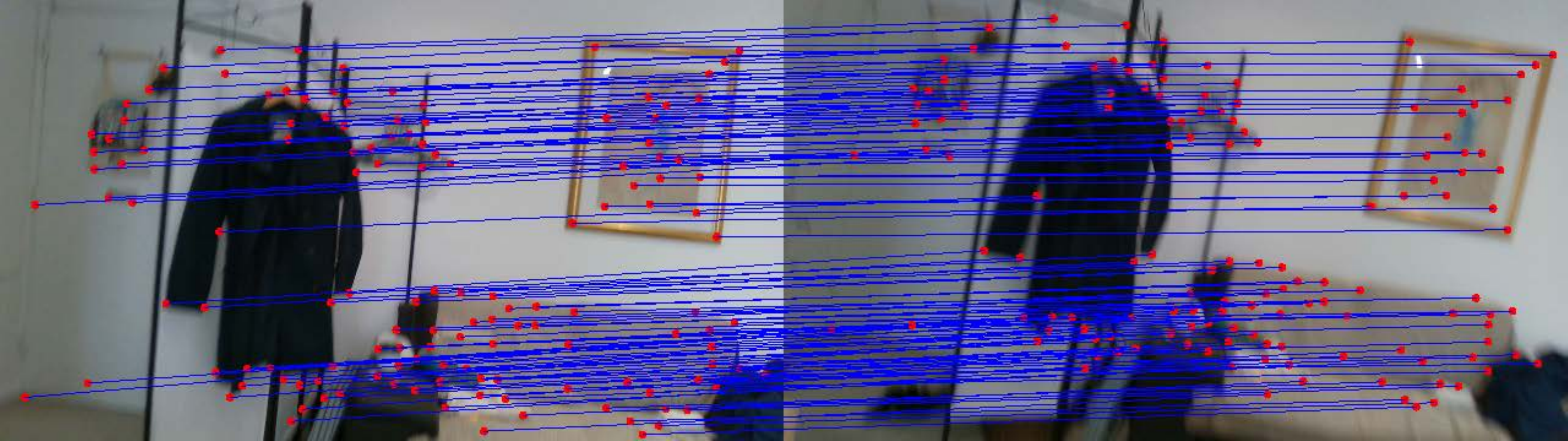}
		\caption{Room}
	\end{subfigure}	 
	\begin{subfigure}[t]{0.33\linewidth}
		\includegraphics[width=\linewidth, height=2cm]{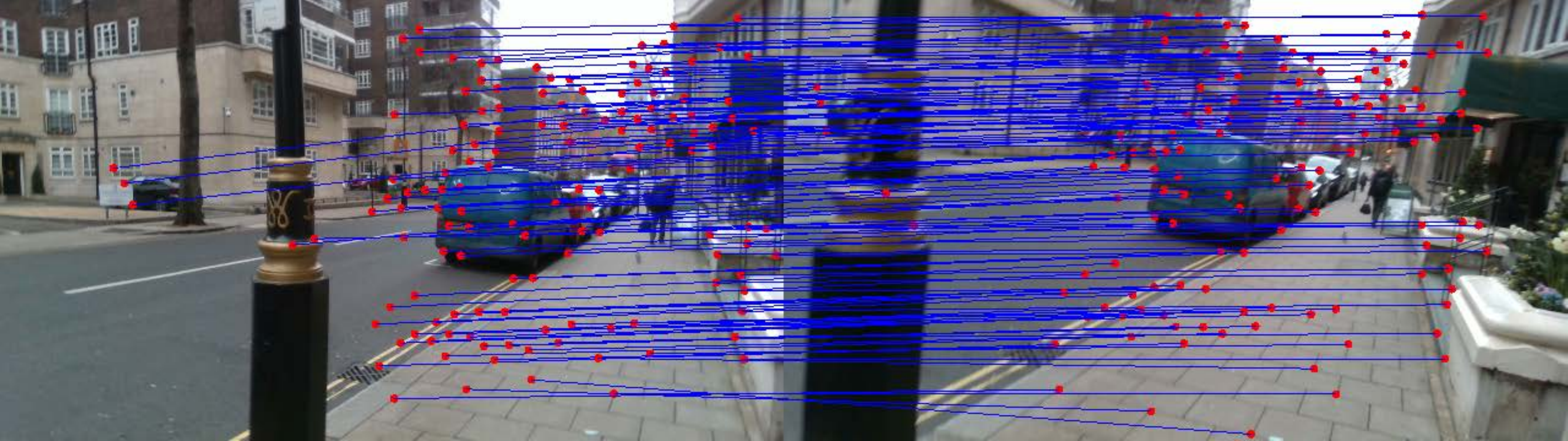}
		\caption{Street}
	\end{subfigure}	 	
	\caption{Sample images and correspondences of the three sequences used in the real data experiments. (Upper row: images, lower row: correspondences) }		
	\label{fig:real_data_correspondences}
	\label{fig:real_data_sample_images} 
	\vspace{-2mm}
\end{figure*}

Then, we perform another experiment on different levels of angular and linear velocities.
The magnitude of the angular velocity increases the same as in the previous experiment and that of the linear velocity increases from 4 (level 1) to 20 $m/s$ (level 5). 
For brevity, we denote the magnitude of the angular and linear velocities using their levels. 
Figure~\ref{fig:exp_synthetic_rs_dw}  shows that G-RSRP, G-RSRP+, and NRSRP are much more accurate than GSRP.
This tendency of the results is similar to that of the previous experiment.
Although the average errors and their standard deviations are a bit increased due to the effects of the linear velocities, the average errors are still less than 2$^{\circ}$ and 10$^{\circ}$ even at the maximum level of the velocities.
It indicates that the effect of the linear velocity is tolerable and not severe. 
Interestingly, G-RSRP and G-RSRP+ are more stable than NRSRP considering angular and linear velocities. 
It shows that even complicated models could sometimes produce incorrect  estimates.

Secondly, we perform other experiments with different levels of noises of  camera and gyroscope measurements in order to investigate the sensitivity of the proposed method. 
For these experiments, the angular velocity is set to 2.5 $rad/s$ and the linear velocity to 20 $m/s$ as in the level 5 of the previous experiments on angular and linear velocities.
We determined to use these large values to clearly see the effects of noises in severe rolling shutter distortions. 
For the experiment on gyroscope noises, we increase the standard deviation of the gyroscope from 0.1 to 0.5 $rad/s$ and set the standard deviation of the camera to 1 $pixel$.
On the contrary, we increase the standard deviation of the camera measurement noise from 1 to 5 $pixel$ and set the standard deviation of the gyroscope to 0.1 $rad/s$. 
Figure~\ref{fig:exp_synthetic_noise} shows that gyroscope noises do not affect on the performance of G-RSRP and G-RSRP+.
However, large camera noises significantly degrade the translation estimation accuracy of G-RSRP.
Interestingly, the proposed nonlinear refinement improves the accuracy of the translation estimation in case that camera noises are severe.
Apart from this, G-RSRP+ shows better performance at various noise levels.
It indicates that the refinements make the estimates stable as we observed in previous experiments on increasing angular velocity.

\subsection{Real Data}

\textbf{Real data preparation:} We evaluate the performance of the proposed method in our datasets collected from an Intel RealSense Camera (ZR300).
ZR300 consists of a rolling shutter camera, a global shutter camera, a depth sensors, and an IMU including a gyroscope and it also provides their synchronized timestamps.
The frame rates of the rolling shutter and global shutter cameras are 30 Hz and that of the gyroscope is 200 Hz. 
We recorded the image sequences of the rolling shutter and global shutter cameras and gyroscope measurements in indoor and outdoor environments with handheld motion. 
Then, we used the images from the global shutter camera to generate ground truth relative poses with GSRP~\cite{Nister:PAMI:2004}.

Figure~\ref{fig:real_data_sample_images} shows the sample images from three sequences that we recorded.
We acquire the $Bench$ and $Room$ sequences for about 1 minute and the $Street$ sequence for about 10 minutes.
In these sequences, we generate image pairs with gyroscope measurements for our relative pose estimation problem.
In addition, since we need image pairs having the enough number of correspondences and large camera motion for our experiments, we basically sample images having large angular velocity motion by examining the gyroscope measurements. 
The sampled images compose image pairs for relative pose estimation. 
We extract many feature points from the images and track them~\cite{opencv_library} across images having large angular motion. 
Figure~\ref{fig:real_data_correspondences} shows the correspondence samples of the generated image pairs.
Although the RealSense Camera provides initial calibration parameters, we perform the calibration~\cite{oth2013rolling,maye2013self,furgale2012continuous,furgale2013unified} of the rolling shutter, global shutter cameras, and the gyroscope by ourselves in order to obtain more accurate calibration parameters.
Since there are no public calibration algorithms on a rolling shutter camera and an IMU, we first perform calibrations of 1) the global shutter camera and the IMU and 2) the global shutter and rolling shutter cameras.
Then, we compute extrinsic parameters between the rolling shutter camera and the gyroscope from the estimated two extrinsic parameters.
Besides, we estimate the intrinsic parameters of the rolling and global shutter cameras, intrinsic parameters of the gyroscope such as biases and the time offset between the gyroscope and cameras. 
Finally, in order to reduce the noises of the gyroscope, we use an average of the gyroscope measurements within 50 $ms$ near the image frame timestamp.

\begin{figure}[t]	
	\centering 	
	\begin{subfigure}[t]{0.99\linewidth}
		\includegraphics[width=\linewidth]{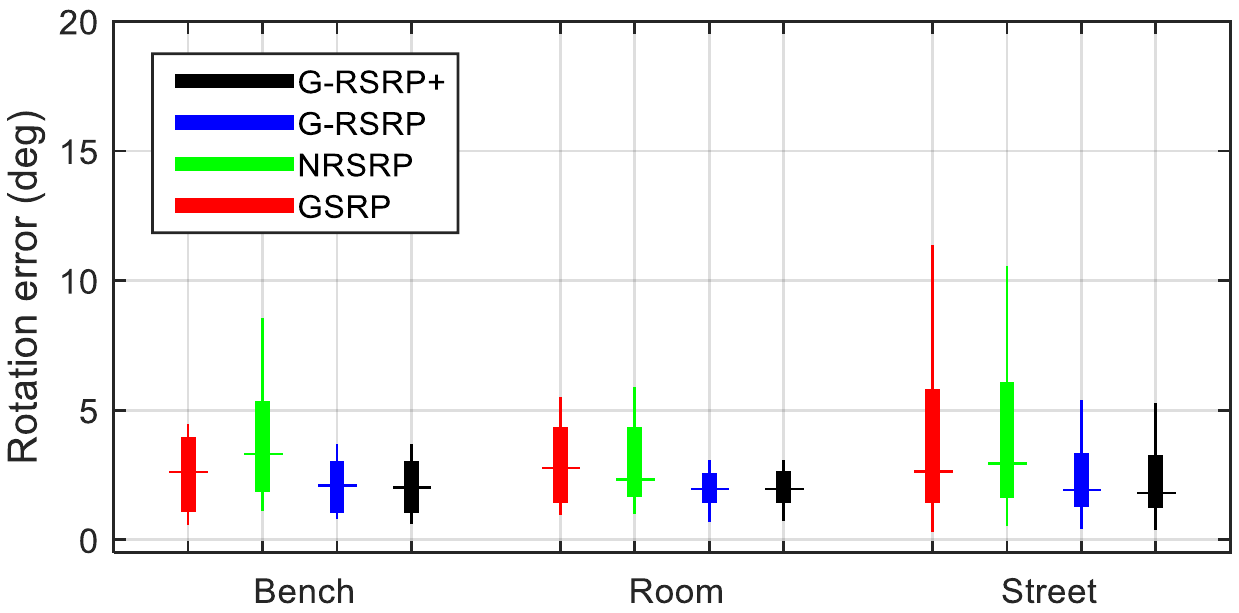}
	\end{subfigure}	
	\begin{subfigure}[t]{0.99\linewidth}
		\includegraphics[width=\linewidth]{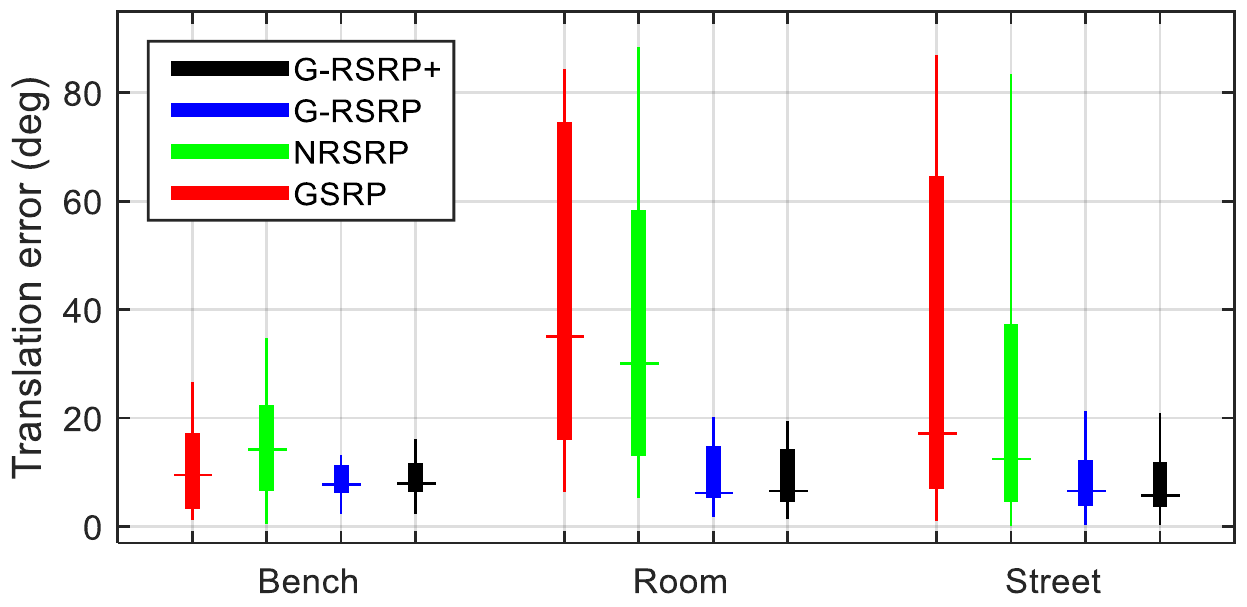}
	\end{subfigure}	 
	\caption{Accuracy comparison for our real datasets}
	\label{fig:real_accuracy_comparison}
\end{figure}

\textbf{Experimental results: }
Figure~\ref{fig:real_accuracy_comparison} shows the performance comparison on three real sequences.
Overall, the rotation and translation errors for all methods are larger than those in the synthetic data experiments for angular and linear velocities.
The reason is that the real dataset contains outliers (incorrect image correspondences)  and noises from the feature tracking on the consecutive frames, and we also have calibration errors.
Besides, there exist degenerate situations on camera motion as pointed out  in~\cite{Albl:ECCV:2016}.
Nevertheless, the proposed G-RSRP and G-RSRP+ outperform GSRP and NRSRP.
In all sequences, the median error and standard deviation of G-RSRP and G-RSRP+ are lower than those of GSRP.
Actually, we can see that G-RSRP and G-RSRP+ produce much less errors than GSRP and NRSRP.
Although NRSRP a bit reduces the standard deviation of the translation errors in $Room$ and $Street$ sequences, it does not make much improvements over GSRP. 
This is because, while the nonlinear optimization of NRSRP depends on the initial value, the initial value obtained from GSRP is erroneous and this lead to little improvement of NRSRP. 
Besides, the complicated model sometimes produces incorrect estimates as we observed in the synthetic data experiments.

\subsection{Time complexity}

Since G-RSRP is the closed-form polynomial solver and the DOF of the problem is small, the computation for obtaining 20 solutions with five points and gyroscope measurements is very fast. 
Besides, it finds inliers in the RANSAC process easily because G-RSRP considers the rolling shutter geometry.
However, GSRP regards rolling shutter distorted points as outliers.
Thus, it can take more time for the RANSAC process.
In general, NRSRP and G-RSRP+ takes much more time because the nonlinear solver estimates a solution in the iterative scheme with a lot of points.

\subsection{Degeneracy}
 
We found that the proposed method fails to estimate the relative pose in planar scenes similarly to the case reported in ~\cite{Albl:CVPR:2016}.
We performed a synthetic data experiment with the 3D landmarks generated in a plane.
Interestingly, we observed that the translation estimates converged to forward motion although the rotation estimates were correct.
This issue can be resolved by homography estimation considering rolling shutter camera geometry. 
We can measure the planarity of the scene (\ie how planar the scene is) via the number of inliers and/or the errors of the measurements. 
Then, we can avoid the degenerate situation with this measure.
The mathematical analysis on degeneracy and homography estimation for rolling shutter cameras sill remain as open problems.
Moreover, instead of a geometric solution such as homography, we can judge whether scenes are planar or not with state-of-the-art machine learning techniques.
Learning the planarity of the scene is another interesting research topic.

\section{Conclusion}

We have proposed a new method to estimate relative pose for a rolling shutter camera with the aid of the gyroscope, angular velocity measurements. 
We exploited angular velocity measurements to simplify the relative pose estimation problem and lowered the DOF of the angular rolling shutter essential matrix from 11 to five. 
Then, we found the minimal solution of the simplified problem using the GB method and refined the result through nonlinear optimization.  
We experimentally verified the proposed method using the synthetic and real data, and confirmed that the proposed method produces accurate relative pose estimates compared to the conventional global shutter and rolling shutter relative pose estimation methods.
The proposed method can be utilized as a essential component for visual-inertial SLAM/SfM with rolling shutter cameras.

{\small
\bibliographystyle{ieee}
\bibliography{irsrp}
}

\end{document}